%% file: main.tex
\documentclass[10pt,twocolumn,letterpaper]{article}

\usepackage{cvpr}              


\usepackage{graphicx}
\usepackage{booktabs}
\usepackage{microtype}
\usepackage[normalem]{ulem}
\usepackage{adjustbox}
\usepackage{multirow}
\usepackage{multicol}
\usepackage{tabularx}
\usepackage{makecell}
\usepackage{etoc}
\usepackage[accsupp]{axessibility}
\usepackage{xspace}
\usepackage{diagbox}
\usepackage{wrapfig}
\usepackage{bbding}
\usepackage{pifont}
\usepackage{wasysym}
\usepackage[accsupp]{axessibility}

\usepackage{url}
\usepackage{algorithm}
\usepackage{algorithmic}
\usepackage[utf8]{inputenc} 
\usepackage[T1]{fontenc}  
\usepackage{url}        
\usepackage{amsfonts}      
\usepackage{nicefrac}   
\usepackage{xcolor}         

\usepackage{amsmath}
\usepackage{color}
\usepackage{colortbl}
\definecolor{cvprblue}{rgb}{0.52,0.20,0.89}
\usepackage[pagebackref,breaklinks,colorlinks,allcolors=cvprblue]{hyperref}

\def\paperID{paper id} 
\def\confName{CVPR}
\def\confYear{2025}

\title{Low-Biased General Annotated Dataset Generation}

\author{
Dengyang Jiang$^{1}$\thanks{Equal contribution.} ~
Haoyu Wang$^{1*}$ ~
Lei Zhang$^{1}$\thanks{Corresponding author} ~
Wei Wei$^{1}$ \\
Guang Dai$^{2}$ ~
Mengmeng Wang$^{3}$ ~
Jingdong Wang$^{4}$ ~
Yanning Zhang$^{1}$ \\[2mm]
\fontsize{10.4pt}{9.84pt}\selectfont
$^{1}$ Northwestern Polytechnical University \hspace{6mm}
$^{2}$ SGIT AI Lab, State Grid Corporation of China \\
\fontsize{10.4pt}{9.84pt}\selectfont
$^{3}$ Zhejiang University of Technology \hspace{6mm}
$^{4}$ Baidu Inc. 
}
    
\input{figtex/begin}

\begin{document}
\maketitle
\begin{abstract}
    Pre-training backbone networks on a general annotated dataset (e.g., ImageNet) that comprises numerous manually collected images with category annotations has proven to be indispensable for enhancing the generalization capacity of downstream visual tasks. However, those manually collected images often exhibit bias, which is non-transferable across either categories or domains, thus causing the model's generalization capacity degeneration. To mitigate this problem, we present a \textbf{l}ow-\textbf{b}iased general annotated dataset \textbf{gen}eration framework (\textbf{lbGen}). Instead of expensive manual collection, we aim at directly generating low-biased images with category annotations. To achieve this goal, we propose to leverage the advantage of a multimodal foundation model (e.g., CLIP), in terms of aligning images in a low-biased semantic space defined by language. Specifically, we develop a bi-level semantic alignment loss, which not only forces all generated images to be consistent with the semantic distribution of all categories belonging to the target dataset in an adversarial learning manner, but also requires each generated image to match the semantic description of its category name. In addition, we further cast an existing image quality scoring model into a quality assurance loss to preserve the quality of the generated image. By leveraging these two loss functions, we can obtain a low-biased image generation model by simply fine-tuning a pre-trained diffusion model using only all category names in the target dataset as input. Experimental results confirm that, compared with the manually labeled dataset or other synthetic datasets, the utilization of our generated low-biased dataset leads to stable generalization capacity enhancement of different backbone networks across various tasks, especially in tasks where the manually labeled samples are scarce. Code is available at: {\url{https://github.com/vvvvvjdy/lbGen}}
\end{abstract}

\input{figtex/introbias}
\section{Introduction}
\label{sec:intro}
Deep neural networks have achieved great success in various computer vision tasks~\cite{seg,detect,transferlearning}. One indispensable premise of such success lies on pre-training the parameter-extensive backbone network using a general annotated dataset (e.g., ImageNet~\cite{imagenet}) that contains a large number of images with manually annotated categories. Profiting from the vast amount of annotated images and the diverse image categories, the pre-trained backbone networks often show pleasing generalization capacity and perform effectively in the target computer vision task through simple fine-tuning it together with a corresponding task head with a few parameters~\cite{maskrcnn,upernet}. Unfortunately, recent studies~\cite{biaskaiming,unbiased,understand_bias} uncover that these manually collected images often exhibit non-travail bias\footnote{In this paper, `dataset bias' refers to `systematic bias introduced in data collection, selection, or processing that impair the generalization capacity of the model'.} (e.g., a certain background, image style, object position for a specific category etc.) which can be easily captured by backbone networks during pre-training, but hardly noticed by human collectors (see Figure~\ref{fig:example}). Such hidden bias is proven to be cast into a shortcut feature representation~\cite{shortcut}  
to improve the in-domain performance but deteriorate the generalization 
capacity of pre-trained backbone networks on target tasks in the cross-category or cross-domain settings~\cite{bias_explain_generalization,noise} which shows an obvious image distribution gap from the utilized general annotated dataset. For example, when a specific category of images often shows a similar background, the pre-train backbone networks will consider the background as the discriminative feature of such a category while overlooking the cross-category or cross-domain transferable semantic features (e.g., shape, structure, etc.). Therefore, it is crucial to obtain a low-biased general annotated dataset to enhance the cross-category or cross-domain generalization capacity of the pre-trained backbones.

To this end, a straight solution is to manually re-collect extensive low-biased images. However, it will not only produce expensive manual collection costs, but also inevitably incur some other undetectable bias. Recently, diffusion models~\cite{imagen,sd1.5,dalle2} have shown powerful capacity in terms of generating high-quality synthetic images based on the text description of image contents, thus providing a feasible way to directly generate images with annotations without manual collection cost. Moreover, some studies~\cite{realfake,RC,fakeit} have demonstrated that those randomly generated images with annotations can be utilized for network training. Although most existing diffusion models can be directly utilized for general annotated dataset generation, they mainly focus on generating images with the distribution consistent with the conventional manually annotated general dataset (e.g., ImageNet) and scarcely attempt to generate low-biased images. Thus, pre-training the backbone networks on these generated general annotated datasets will not bring non-travail generalization capacity enhancement~\cite{benchmarkbias}.  

To mitigate this problem, we present a \textbf{l}ow-\textbf{b}iased general annotated dataset \textbf{gen}eration framework (\textbf{lbGen}), which takes the first attempt to directly generating synthetic low-biased images with category annotations. To achieve this goal, we first have to define a low-biased space where the feature representation of each image emphasizes transferable semantic characteristics. Considering that the observation that text information is closer to the ideal semantic information and the recent progress of multimodal foundation model (e.g., CLIP~\cite{clip}) which aims at mapping images into a low-biased semantic space defined by language, a straight idea is to constrain the image output of the existing diffusion models to follow the semantic distribution of the specific image category in such a low-biased semantic space. Following this idea, we develop a bi-level semantic alignment loss based on the CLIP model to fine-tune the pre-trained diffusion model. In a specific, on the one hand, such a loss forces all generated images to be consistent with the semantic distribution of all categories belonging to the target dataset in the CLIP feature space using an adversarial learning scheme. On the other hand, it also requires each generated image to match the semantic description of its category name in the CLIP feature space using a simple cosine similarity metric. By doing these, we can obtain a low-biased image generation model by simply fine-tuning a pre-trained diffusion model using only all category names in the target dataset as input. In addition, to sidestep image quality degradation caused by the low-biased image generation learning, we further cast an existing image quality scoring model into a quality assurance loss to assist the bi-level semantic alignment loss for diffusion model fine-tuning. 

To testify the efficacy of the proposed framework, we pre-train two conventional backbone networks on our generated low-biased general annotated dataset and training specific heads on different downstream tasks. Compared with backbone networks pre-trained either on the manually collected generated dataset or that generated by existing diffusion models, our approach achieves obvious generalization performance improvement, especially when the manually annotated samples in the target task are scarce. Moreover, additional experiments prove that our pre-trained backbone networks capture lower specific bias (e.g. context, background, shape-texture), which further demonstrates the generality of our framework.  

In summary, our main contributions are as follows:

\begin{itemize}		
		\item We propose the first low-biased general annotated dataset generation framework, which jumps out of the dilemma of traditional manual data collection in terms of mitigating dataset bias.
		\item We present a bi-level semantic alignment module assisted by a quality assurance loss to simply fine-tune the standard diffusion model using only all category names in the target dataset as input.
		\item With our generated general low-biased dataset, the pre-trained backbone network shows state-of-the-art generalization capacity in different downstream tasks.  
\end{itemize}
	
\section{Related Work}
\subsection{Datasets Bias}
Since the deep learning revolution in 2012~\cite{alexnet}, the large-scale manually collected annotated dataset (e.g., ImageNet)  performs no longer a simple training dataset for its own tasks, but a general dataset utilized for backbone network pre-training which  has become the indelible step for enhancing the generalization performance of various downstream tasks.  However, recent studies~\cite{biaskaiming,unbiased,understand_bias} have consecutively revealed that these existing manually collected general dataset exhibit non-trial bias, which results in sup-optimal cross-categories and cross-domain generalization capacity, especially when the manually annotated samples in the target task are scarce. For example, Liu~\etal~\cite{biaskaiming} observe that deep neural networks can achieve excellent accuracy in classifying which dataset an image is from. In other words, the neural network discovers some dataset-specific patterns, a form of bias. In addition, studies in~\cite{benchmarkbias,ood1,transferlearning} attempt to implicitly measure the dataset bias by investigating the cross-category or cross-domain generalization capacity as well as the robustness of the models pre-trained on the dataset. 

However, these works mainly focus on raising the dataset bias problem or bias measurement. In contrast, in this study, we take the first attempt to solve this problem and aims at borrowing the advantage of diffusion model in image generation to directly generate a low-biased general dataset for better backbone pre-training.  

\input{figtex/method}
\subsection{Synthetic Dataset Generation}
Different from manual collection, synthetic dataset generation aims at directly generating image using deep neural networks based on some text description. For example, in early days, Zhu~\etal~\cite{zhu2019dm, xu2018attngan} utilized adversarial generative networks to model the mapping relationship between the input text description and the output image. However, these methods require large-scale high-quality images from target categories for network training, and the generalization capacity to unknown text descriptions is limited. More recently, as diffusion model shows more powerful capacity in generalization to unknown text description~\cite{sd1.5,sdxl,sd3}, some works have attempted to utilize the diffusion model to generate ImageNet-like synthetic dataset for backbone pre-training. For example, Bansal~\etal~\cite{RC} 
directly fine-tune the diffusion model on ImagNet-1K~\cite{imagenet} and use meticulously designed prompts to generate the dataset. Lei~\etal~\cite{cip} utilizes ViT-GPT2~\cite{vitgpt2} to get a unique prompt to generate each image. Yuan~\etal~\cite{realfake} resort to learn the distribution of ImageNet and use Blip-captions~\cite{blip2} of ImageNet as prompts to synthesize the dataset. 

Although these diffusion-based methods can be utilized for general dataset generation, they mainly focus on simulating the existing ImageNet without considering the dataset bias. 
In this study, we attempt to fine-tune the diffusion model to directly generate a low-biased general annotated dataset without using any image-text pairs but the category names of the target dataset as input. 

\section{Approach}
\label{sec:appro}

The overall training framework of lbGen is shown in Figure~\ref{fig:method}. In Section~\ref{pre}, we begin with elucidating the basic methodology employed in our training process. In Section~\ref{method:s_alignment}, we then illustrate the bi-level semantic alignment module, which is the core part of our approach. Subsequently, in Section~\ref{method:q_assurance}, we further introduce the quality assurance module for fidelity preserving the images, and finally we integrate the two components for joint learning.

\subsection{Preliminary}
\label{pre}
We implement our method on the leading text-to-image diffusion model, Stable Diffusion~\cite{sd1.5}, which belongs to the family of latent diffusion models (LDM)~\cite{ldm}. In the traditional training process, a normally distributed noise $\epsilon$ is added to the original latent code $z_0$ with a variable extent based on a timestep $t$ sampling from $\{1,..., T\}$. Then, a denoising function $\epsilon_\theta$, parameterized by a UNet backbone, is trained to predict the noise added to $z_0$ with the text prompt $\mathcal{P}_{t}$ and the current latent $z_t$ as the input. 
Specifically, the text prompt is first encoded by a text encoder $W$, then incorporated into the denoising function $\epsilon_\theta$ by the cross-attention mechanism. The denoising loss in diffusion models' training is formally expressed as:
\begin{equation}
\mathcal{L}_{\text {LDM}}=\mathbb{E}_{z_0, t, p, \epsilon \sim \mathcal{N}({0}, {I})}\left[\left\|{\epsilon}-{\epsilon}_{{\theta}}\left(z_t, t, W(\mathcal{P}_{t})\right)\right\|^2\right]. \label{eq:lossldm}
\end{equation}
For inference, the process can be formulated as a Markov decision process that iteratively estimates the noise and computes the next latent sample: \begin{equation}
p_\theta(z_0|W(\mathcal{P}_{t})) = p(z_t)\prod_{t=1}^T p_\theta(z_{t-1}|z_t, W(\mathcal{P}_{t})). \label{denoise} 
\end{equation}

However, this method requires vast quantities of image-text pairs for training and it takes extended times to converge, which is not feasible to fine-tune a model under our restrictive conditions. To overcome these difficulties, we follow some works~\cite{comat,text_re,video_re} to fine-tune the diffusion model with reinforcement learning (RL). Different from the original loss function in \cref{eq:lossldm}, given a reward function $R$(.), the objective of RL is to maximize the expected reward:
\begin{equation}
    J(\phi) = \mathbb{E} \left[R(Z_0, c \right]. \label{eq:rl_objective}
\end{equation}
With the denoising process showed in \cref{denoise} , the gradient when fine-tuning diffusion models with reward feedback \cref{eq:rl_objective} can be computed as:
\begin{equation}
    \nabla_\phi J = \mathbb{E}\left[ \sum_{t=0}^T \nabla_\phi \log p_\theta(z_{t-1}|z_t, W(\mathcal{P}_{t}) R(Z_0, c) \right]. \label{eq:rl_gradient}
\end{equation} Noticing that only text prompt $\mathcal{P}_{t}$ and conditional context $c$ for reward models are required in this training paradigm instead of a dataset containing image-text pairs, which is consistent with the fact that we do not incorporate any external biased images in our training process.

\subsection{Bi-Level Semantic Alignment}
\label{method:s_alignment} 
As mentioned in Section~\ref{sec:intro}, we assume that the semantic space defined by language can be a low-biased representation and our key insight is using this characteristic of language to fine-tune a pre-trained diffusion model as our lbGen generator. We achieve this by leveraging a simple Linear-ReLU-Linear based discriminator $\mathcal{D}_\phi$ and utilizing CLIP to carry out a bi-level semantic alignment. We use only 1000 class names of ImageNet as our inputs, which ensures that no other biased information is introduced except semantic information of the dataset.

\noindent\textbf{Entire Dataset Alignment.} Building on the advantages of CLIP which has unified the image and the text into one represention space, we can initially use CLIP text encoder to extract text features $\{f_{c_1}, f_{c_2}, \dots ,f_{c_{1000}}\}$ of classnames $\{c_1, c_2, \dots ,c_{1000}\}$. We consider these text features to be a low-biased semantic distribution of the entire ImageNet. Then, we generate an image using prompt $c_i$ and send it to CLIP image encoder to get the image feature $f_{im_i}$. Next, we randomly choose a text feature $f_{c_j}$ from $\{f_{c_1}, f_{c_2}, \dots ,f_{c_{1000}}\}$. It is worth emphasizing that we do not use the text feature which belongs to the same class as the image feature since we aim to align the whole synthetic dataset to its general semantic representation space regardless of the concrete class.  Finally, features $f_{im_i}$ and $f_{c_j}$ are fed into $\mathcal{D}_\phi$ for computing entire semantic alignment loss as follows:
\begin{equation}
\mathcal{L}_{en} = \log \left(\mathcal{D}_\phi\left(f_{c_j}\right)\right) + \log \left(1-\mathcal{D}_\phi\left(f_{im_i}\right)\right).
\end{equation}
Similar to the training object of Generative Adversarial Network(GAN)~\cite{gan}, we expect to fine-tune the diffusion model to minimize this adversarial loss, while concurrently training the discriminator to maximize it.

\noindent\textbf{Individual Image Alignment.} Expect for mapping the images to be consistent with the semantic distribution of all classes within the entire dataset, we need to precisely control each category of images to match their semantic description. To this end, we introduce the individual semantic alignment loss.
In particular, given the generated image ${im_i}$ using class name $c_i$, we use simple \texttt{"photo of $c_i$"} as the low-biased semantic description $p_{c_i}$ and send them to CLIP. Different from using CLIP to align the dataset globally in the entire semantic space, we aim at forcing the semantic information of each image to be dovetailed with its class by maximizing the cosine similarity between the image and its corresponding semantic description through CLIP. Thus, we can obtain $\mathcal{L}_{in}$ formulated as follows:
\begin{equation}
\mathcal{L}_{in} = 1 - \frac{f_{im_i} \cdot f_{p_{c_i}}}{\Vert f_{im_i}\Vert \cdot \Vert f_{p_{c_i}}\Vert},
\end{equation}
where $f_{im_i}$ and $f_{p_{c_i}}$ represent the image and text feature vectors extracted by CLIP, the dot product of the vectors is denoted by $\cdot$, and $\Vert\cdot\Vert$ denotes the norm of the vectors. 

By considering these two distinct levels of losses, we can finally add them together and obtain bi-level semantic alignment loss $\mathcal{L}_{bi}$ to refine the diffusion model to align more closely with the low-biased semantic reference.

\subsection{Quality Assurance}
 In practice, only under supervision in terms of text semantics, we observe that the quality of the generated images is sub-optimal after training. Thus, we introduce the quality assurance loss to assist the bi-level semantic alignment loss.

To be specific, we use the state-of-the-art image quality scoring model Q-ALIGN~\cite{qalign} as our quality assurance model. After fine-tuning a Leading open-source multimodal large language model (MLLM) mPLUG-Owl-2~\cite{mplug2} on carefully collected image quality assessment datasets, Q-ALIGN can achieve satisfactory image quality scoring performance. Feeding the generated image ${im_i}$ with the system prompt \texttt{"How would you rate the quality of this image?"} into Q-ALIGN, we can obtain the quality score $Q(im_i)$, which ranges in $[1, 5]$, of the image and using it to calculate quality assurance loss ($\mathcal{L}_{q}$) for diffusion model to optimize. Details about how Q-ALIGN scores images can be found in Appendix.
\begin{equation}
\mathcal{L}_{q} = 1 - \frac{Q(im_i)}{5}.
\end{equation}

Finally, we combine the losses in the bi-level semantic alignment module and quality assurance module to build up our final training object for lbGen generator as follows:
\begin{equation}
    \mathcal{L}  =   \mathcal{L}_{bi} + \lambda_1 \mathcal{L}_{q},
\end{equation}
where $\lambda_1$ is a scaling factor to balance the losses. The pseudocode of the integrated loss computation process can be found in Appendix. 
\label{method:q_assurance} 

\input{tabel/transfer}

\section{Experiment}

\subsection{Experimental Settings}
\noindent\textbf{Datasets.} We choose two recent open-source synthetic ImageNet (GenRobust~\cite{RC}, RealFake~\cite{realfake}) and ImageNet-1K~\cite{imagenet} for comparison. We test the generalization ability and robustness of the pre-trained model using eight transfer learning datasets (Aircraft~\cite{aircraft}, Cars196~\cite{cars}, DTD\cite{dtd}, EuroSAT\cite{eurosat}, Flowers\cite{flowers}, Pets\cite{pets},
Food101\cite{food101}, SUN397\cite{sun}), two visual perception datasets (COCO~\cite{coco}, ADE20K~\cite{ade20k}), as well as three specific bias measurement datasets (FOCUS\cite{focus}, Mixed-Rand \& Mixed-Same~\cite{background}, Cue Conflict~\cite{cueconf})

\noindent\textbf{Implementation Details.} We implement our method on SD1.5~\cite{sd1.5} and fine-tune it with LoRA~\cite{lora}. We choose openai-CLIP-VIT-L~\cite{clip} as our default CLIP model. For the visual backbones, we choose two representative models, ConvNets-based ResNet50~\cite{resnet} and Transformer-based ViT-S~\cite{vit}. During fine-tuning the generator,  we follow Deep Reward \cite{deepreward} and CoMat~\cite{comat} that only enable gradients in 5 steps out of those 50 steps to save GPU memory. During training visual backbones, we maintain the same training hyperparameters across all selected datasets to make a fair comparison.

More details about training hypersettings, data synthesis, datasets for evaluation, and computing resources are provided in Appendix.

\subsection{Generalization across Downstream Tasks}
\noindent\textbf{Transfer Learning.} Transfer learning~\cite{transferlearning} is a widely known downstream visual task and can be significantly influenced by the generalization of the pre-trained model. In our work, we aim to indicate whether the utilization of our lbGen data can enable the backbones to learn better  transferable patterns. To this end, we follow fakeit~\cite{fakeit}, which uses pre-trained visual backbones as feature extractors and train simple linear logistic regression classifiers from scratch.

\input{figtex/tend}
\input{tabel/seg_det}
As illustrated in Table~\ref{tab:transfer}, we observe that models pre-trained on our dataset outperform all other candidates. Compared with the second-best synthetic datasets, we achieve $+1.4\%$ and $+3.6\%$ leading performance on ResNet50 and ViT-S respectively. More importantly, our method exhibits $1.7\%$ and $2.1\%$ average accuracy improvement and superior results in the vast majority of transfer learning datasets compared to real data. Furthermore, we investigate the transfer ability of models pre-trained on real images and our generated images when using less transfer learning training data. Such few-shot setting~\cite{few-shot} requires an even higher generalization capacity of the model. As shown in Figure~\ref{fig:tend}, it is important to note that the advantage of pre-trained models on our data is even greater when there are fewer downstream images for training. This phenomenon further underscores the lower bias of our data and the stronger generalization of the resulting model.
Meanwhile, another striking finding is that achieving high accuracy on the ImageNet validation set does not necessarily correlate with enhanced cross-category generalization performance, thereby enabling us to draw more definitive conclusions regarding the dualistic impact of bias in the existing dataset on the model.

\noindent\textbf{Visual Perception.} Detection and segmentation are two of the most popular downstream visual perception tasks, at the same time, these two tasks can benefit a lot from a well pre-trained backbone. Hence, we want to find out if the utilization of our lbGen data can also enhance the performance on these tasks. To this end, we follow previous studies~\cite{swin,convnext,transnext}, which use Mask R-CNN~\cite{maskrcnn} as detection head for COCO object detection and UperNet~\cite{upernet} as segmentation head for ADE20K semantic segmentation, to evaluate the performance of pre-trained ResNet50 backbone on different datasets. Moreover, to thoroughly test the generalization ability of the backbone, we progressively decrease the number of training samples and observe the outcomes.

The experimental results in Table \ref{tab:seg_det} demonstrate the effectiveness of our lbGen datasets. Although pre-training on real data achieves marginally better performance with full training data, the performance of lbGen pre-trained model consistently outperforms all other pre-training data when downstream data is limited. Specifically, with only $20\%$ of the original training data, the model achieves the highest performance across both tasks, showing gains of up to $1.54\%$ AP$^{box}$ and $1.47\%$ mIoU compared to that pre-trained on IN-Real. This result is particularly valuable for real-world applications where collecting and annotating task-specific training data is often costly and time-consuming. 

\subsection{Robustness Against Specific Bias}
In this section, we aim to figure out whether our lbGen data can help the backbones to learn good features instead of capturing specific bias as a shortcut. Hence, we follow one recent study~\cite{benchmarkbias} to test shape-texture bias, context bias, and background bias of the backbone networks. All these results are given in Table~\ref{tab:specific bias}.

\noindent\textbf{Shape-Texture Bias.} Prior work shows humans primarily use shape for object recognition~\cite{shapebias,shapebias2}, while neural networks often rely on texture cues~\cite{bias_explain_generalization,cueconf}. Hence, we evaluate whether our data can reduce texture bias using the Cue Conflict, where shape and texture cues intentionally conflict across 1200 images from 16 classes. We use $TI$ which represents the texture inclination of the model to understand the decision-making of the model when facing a shape-texture conflicting image (e.g. a cat with the texture of an elephant).
Our findings indicate that training on our lbGen images, models tend to be less texture-biased. Concretely speaking, the two types of models trained on our data show $4.8\%$ and $9.8\%$ texture inclination decline compared with those trained on real images.

\input{tabel/specificbias}
\noindent\textbf{Context Bias.} Context bias means that a model is biased towards using context cues to classify objects rather than learning real object appearance. In the Focus which we use to evaluate the context bias, each image is annotated with the object class, the time of day, location, and weather labels. These images are divided into common and uncommon sets. Uncommon samples are uncommon contexts like “airplane in the forest”. Then we use mutually exclusive partitions of this dataset $P_k$ where $k$ is the number of uncommon attributes and report $CB_{avg.}$ metrics, which is defined as the average relative accuracy between the accuracy on the partition with no uncommon attributes $P_0$ and a partition with k uncommon attributes when k changes from $1$ to $3$: 
\begin{equation}
    CB_{avg.} = \frac{1}{3} \times \sum_{k=1}^3 \frac{Acc_{P_k}}{Acc_{P_0}}.
\end{equation}
In our evaluation result, we find the models trained on our lbGen ImageNet demonstrate leading object recognition capabilities where achieve $64.7\%$ and $66.0\%$ average relative accuracy on each backbone in a constantly changing context.

\input{tabel/ablation}
\noindent\textbf{Background Bias.} The background bias of models can be
used to identify if the model is using the background of the image during training to improve the classification accuracy instead of using the object itself. For the two datasets that we utilize to evaluate the background bias, the Mixed-Rand segments the foreground object in an image and switches the original background with a random background from a different class label, while the Mixed-Same partition places the segmented foreground object on a random background from the same class label. Thus, we can use $BG_{Gap}$ which measures the difference in performance between these two datasets to examine how decision-making processes can be influenced just by changing the background to a different class.
As we report in Table~\ref{tab:specific bias}, The ResNet50 and ViT-S trained on our data obtain $6.4\%$ and $6.1\%$ performance gaps, which shows lower gaps compared with training on other synthetic data and real data. 

\input{figtex/loss-in}
\subsection{Ablation Study}
In this section, we investigate the design choices of lbGen training process. Due to computational cost, without loss of generality, we conducted the ablation study on the smaller ImageNet-100 datasets~\cite{imagenet100} for evaluation. Unless otherwise specified, we choose ResNet50 backbone and mainly report the results of transfer learning. 

\noindent\textbf{Effect of Bi-Level Semantic Alignment Loss.} According to Table~\ref{tab:ablation}, it is notable that without the entire semantic alignment part, the average accuracy decreased by $4.9\%$. Furthermore, when we remove individual semantic alignment part, the accuracy on IN100-val and transfer learning data shows a significant decline. In our analysis, it appears that the absence of individual semantic alignment leads the model to solely learn the overall semantic distribution of the dataset, resulting in a lack of distinction or specific semantic meaning among different classes (see the left picture of each pair in Figure~\ref{fig:loss_in}). This ultimately causes a collapse when training the backbone. In sum, the bi-level semantic alignment loss successfully help to align the generated images into low-biased semantic space from both the entire dataset distribution level and the specific object category level. 

\noindent\textbf{Effect of Quality Assurance Loss.} As displayed in Figure~\ref{fig:loss_q}, adding quality assurance loss sufficiently solves the quality deterioration when only using bi-level semantic alignment loss. Moreover,  results in Table~\ref{tab:ablation} indicate that the accuracy on both ImageNet100 and transfer learning tasks could decline in the absence of quality assurance loss. Thus, it can be drawn that the quality assurance loss successfully help to guarantee the low-level image quality during generation. 

\input{figtex/loss-q}
\noindent\textbf{Effect of CLIP's Capacity.} As depicted in Table~\ref{tab:ablation}, we aim to explore the effect of the knowledge of CLIP model as it plays a pivotal role in our method. Results indicate that the capacity of CLIP matters, the average accuracy decreased to $56.7\%$ when changing to a smaller size which contains lower image-text alignment capacity than a larger one.
\label{sec:results}

\section{Conclusion}
In this study, we take the first attempt to directly generate a low-biased annotated dataset for more generalized backbone network pre-training. Specifically, we develop a novel bi-level semantic alignment loss, which not only forces all generated images to be consistent with the semantic distribution of all categories belonging to the target dataset, but also requires each generated image to match the semantic description of its category. Through fine-tuning the pre-trained diffusion model with the proposed loss together with a quality assurance loss which helps to guarantee the low-level image
quality, we can obtain a low-biased annotated dataset generation model using only all category names in the target dataset as input. Experiments on various  tasks demonstrate that pre-training backbone network on our generated dataset can lead to stable generalization capacity enhancement.

\section{Acknowledgment}
This work is supported in part by the National Natural Science Foundation of China under Grand 62372379, and Grant 62472359; in part by the Xi’an’s Key Industrial Chain Core Technology Breakthrough Project: AI Core Technology Breakthrough under Grand 23ZDCYJSGG0003-2023.

{
    \small
    \bibliographystyle{ieeenat_fullname}
    \bibliography{main}
}

\clearpage
\appendix
\input{sup/sup_arxiv}
\end{document}

%% file: figtex/begin.tex
\makeatletter
\g@addto@macro\@maketitle{
\vspace{-1.5em}
\begin{figure}[H]
\setlength{\linewidth}{\textwidth}
\setlength{\hsize}{\textwidth}
\centering
\includegraphics[width=\textwidth]{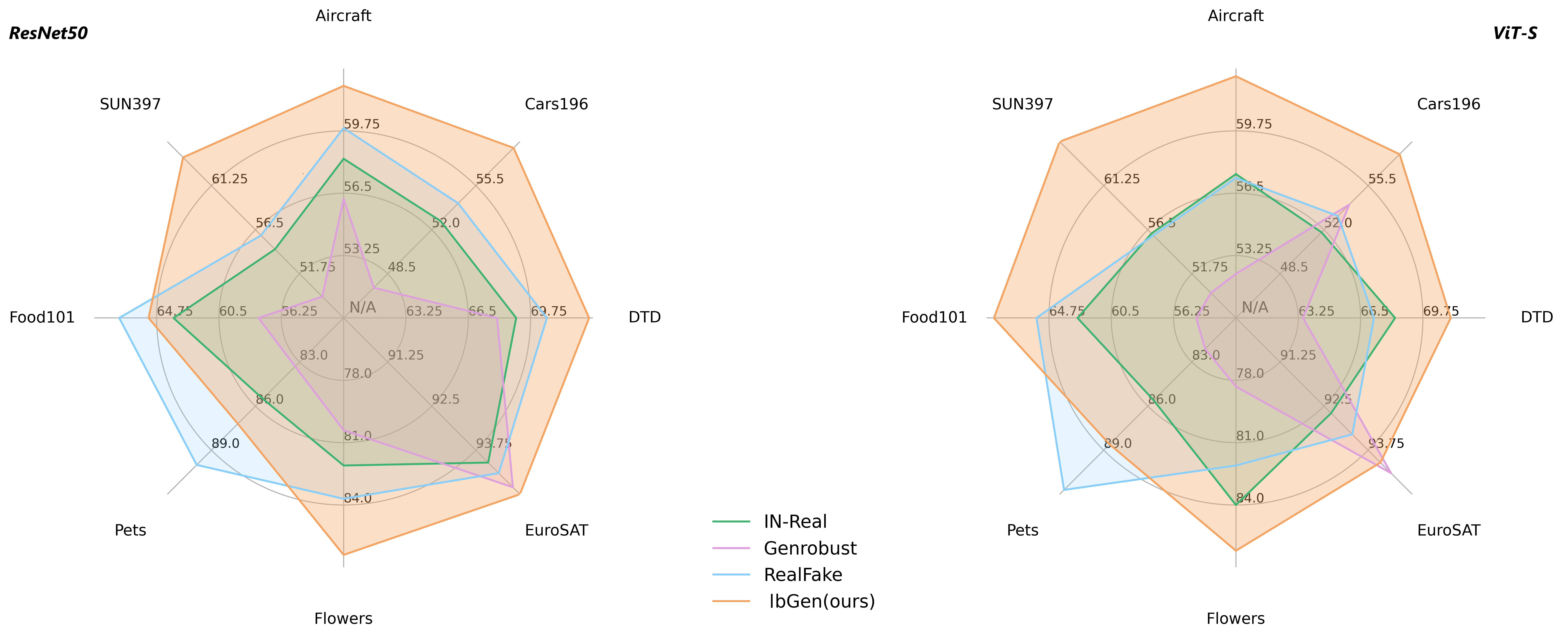} 
\caption{\textbf{Performance on eight transfer learning datasets.}  Pre-training on our generated low-biased general annotated dataset can bring stable generalization capacity enhancement of different backbones.}

\label{fig:radar}
\end{figure}
}
\makeatother

%% file: figtex/introbias.tex
\begin{figure*}[t]
		\centering
        \vspace{-1.0em}
		\includegraphics[width=0.88\linewidth]{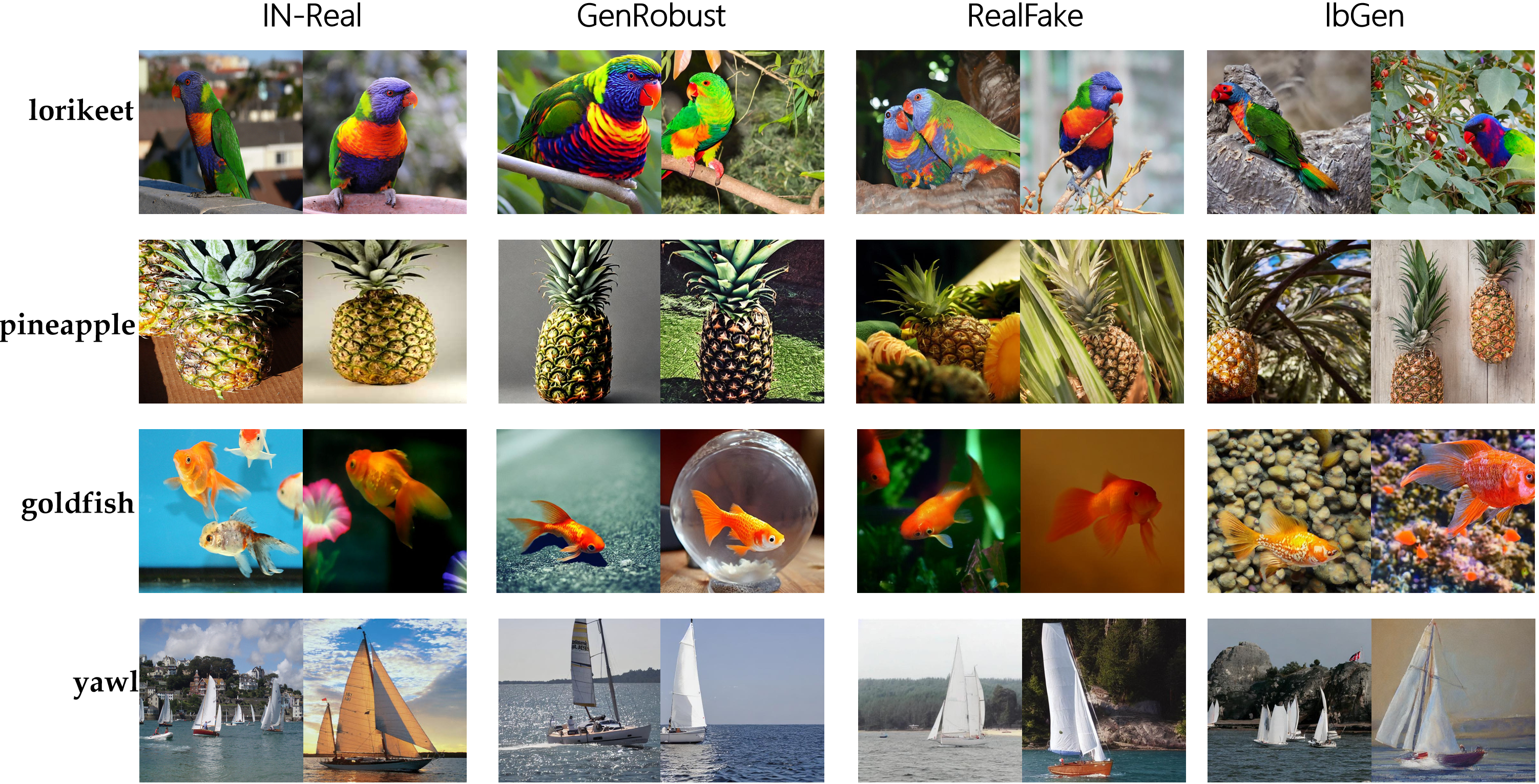}
		\caption{\textbf{Visualization of some randomly sampled images from 4 datasets.} It is hard to tell from which dataset exhibits low bias through these images. However, models trained on these four datasets demonstrate a significant disparity in their generalization capabilities.}
        \vspace{-1em}
		\label{fig:example}
	\end{figure*}

%% file: figtex/method.tex
\begin{figure*}[t]
    \centering
    \vspace{-1.5em}
    \includegraphics[width=1\linewidth]{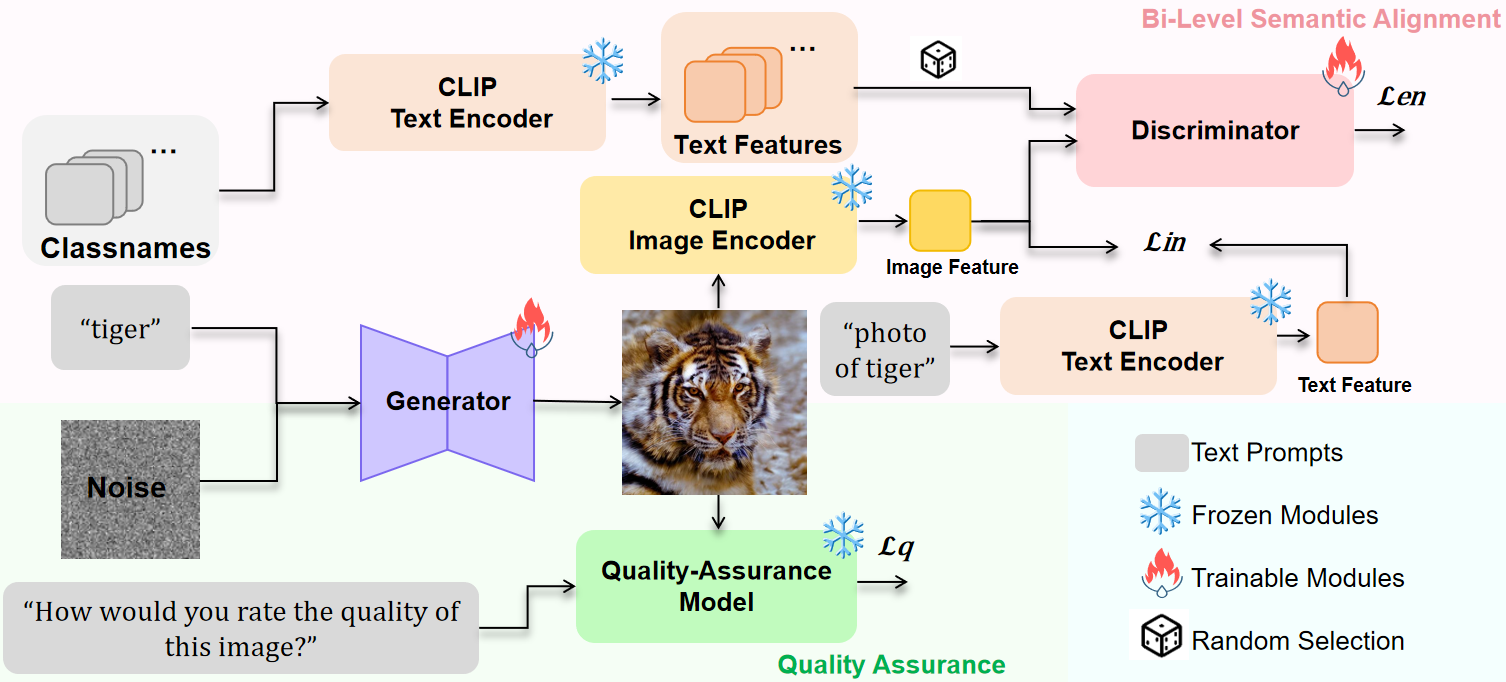}
    \caption{\textbf{Overview of our training method.} The generator first generates an image according to the class name. Then the image is sent to bi-level semantic guidance module and quality assurance module respectively for loss calculation.}
    \label{fig:method}
    \vspace{-1em}
\end{figure*}

%% file: tabel/transfer.tex
\begin{table*}[t]
    \centering
    \vspace{-1.5em}
    \resizebox{1.0\linewidth}{!}{\begin{tabular}{ll|c|ccccccccc}
    \toprule
    Backbone & Pre-Trained Data & IN-val & Aircraft & Cars196 & DTD & EuroSAT & Flowers & Pets & Food101 & SUN397 & Avg.\\
    \specialrule{2pt}{3pt}{5pt}
    \multirow{5}{*}{ResNet50} & IN-Real~\cite{imagenet} & $76.2$ & $56.0$ & $51.5$ & $70.0$ & $93.7$ & $81.4$ & \textcolor{red}{$90.7$} & \underline{$67.2$} & \underline{$56.8$} & $71.5$ \\
    & IN-SD1.5~\cite{sd1.5} & $45.8$ & $58.3$ & $52.7$ & $69.0$ & $94.1$ & $82.1$ & $85.5$ & $63.6$ & $54.4$ &$67.3$ \\ 
    & IN-GenRobust~\cite{RC} & $43.4$ & $56.2$ & $47.4$ & $68.0$ & \underline{$94.8$} & $80.4$ & $83.3$ & $57.8$ & $49.3$ & $64.5$ \\ 
    & IN-RealFake~\cite{realfake} & $69.8$ & \underline{$59.9$} & \underline{$54.1$} & \underline{$70.6$} & $94.4$ & \underline{$83.7$} & \underline{$90.0$} & \textcolor{red}{$67.3$} & $55.9$ & \underline{$71.8$} \\ 
    \rowcolor{Gray!16}
    & IN-lbGen(ours) & $46.1$ & \textcolor{red}{$62.1$} & \textcolor{red}{$58.5$} & \textcolor{red}{$72.8$} & \textcolor{red}{$95.0$} & \textcolor{red}{$86.4$} & $87.2$ & $65.3$ & \textcolor{red}{$64.3$} & \textcolor{red}{$73.2$} \\
    
    \specialrule{2pt}{3pt}{5pt}
    \multirow{5}{*}{ViT-S} & IN-Real~\cite{imagenet} & $78.7$ & \underline{$59.4$} & \underline{$56.4$} & \underline{$69.5$} & \underline{$94.1$} & $83.0$ & \underline{$90.2$} & \underline{$68.3$} & \underline{$57.2$} & \underline{$72.3$} \\
    & IN-SD1.5~\cite{sd1.5} & $46.6$ & $57.5$ & $51.8$ & $68.3$ & $92.7$ & \underline{$84.0$} & $85.6$ & $62.8$ & $56.1$ & $69.9$ \\
    & IN-GenRobust~\cite{RC} & $44.9$ & $52.3$ & $54.0$ & $63.5$ & \textcolor{red}{$94.4$} & $78.3$ & $82.1$ & $54.7$ & $49.7$ & $66.3$ \\
    & IN-RealFake~\cite{realfake} & $72.3$ & $57.3$ & $53.1$ & $67.2$ & $93.3$ & $82.1$ & \textcolor{red}{$91.7$} & $65.6$ & $55.9$ & $70.8$ \\
    \rowcolor{Gray!16}
    & IN-lbGen(ours) & $46.3$ & \textcolor{red}{$62.6$} & \textcolor{red}{$58.0$} & \textcolor{red}{$71.2$} & \underline{$94.1$} & \textcolor{red}{$86.2$} & $88.6$ & \textcolor{red}{$68.5$} & \textcolor{red}{$66.0$} & \textcolor{red}{$74.4$} \\
    
    \bottomrule
    \end{tabular}}
    \caption{
        \textbf{Top-1 accuracy on transfer learning datasets.} The average accuracy across eight transfer learning datasets is denoted as Avg. The \textcolor{red}{best} and \underline{second-best} transfer learning performance of each backbone are highlighted in red and underlined. IN-SD1.5 denotes only using original SD1.5 to generate the data. We also present results on ImageNet validation set for reference.}
    \label{tab:transfer}
    \vspace{-1em}
\end{table*}

%% file: figtex/tend.tex
\begin{figure}[t]
    \centering    
    \includegraphics[width=\linewidth]{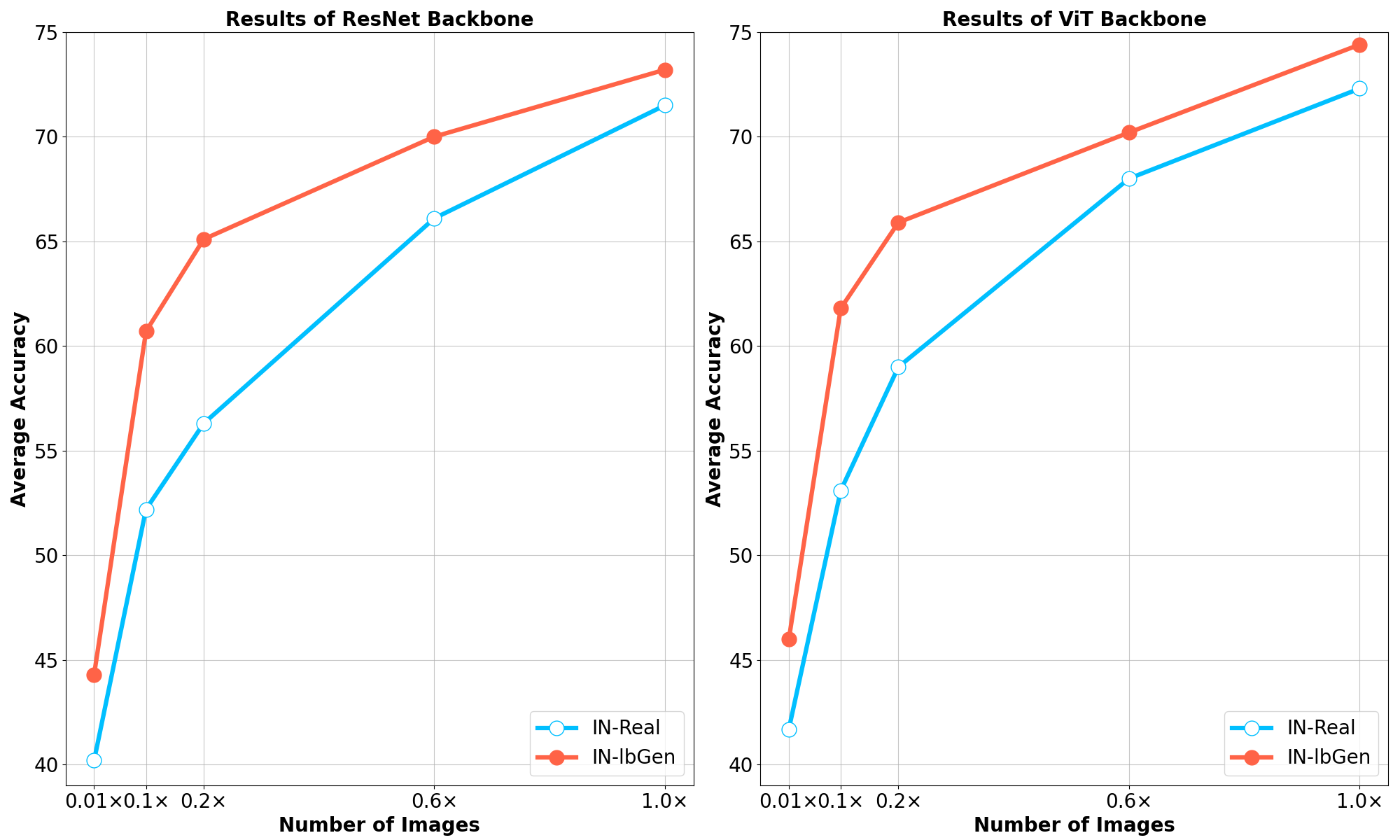}
    \caption{\textbf{Scaling down the number of training images of eight transfer learning datasets.} The benefits of using pre-trained models on our lbGen images are even more pronounced when there is less data for training.}
    \label{fig:tend}
    \vspace{-1em}
\end{figure}

%% file: tabel/seg_det.tex
\begin{table}[t]
    \centering
    \begin{subtable}[t]{1.0\linewidth}
        \centering
        \resizebox{0.86\linewidth}{!}{%
        \begin{tabular}{l|cccc}
        \toprule
         \multirow{2}{*}{Pre-Trained Data} & \multicolumn{4}{c}{COCO ($AP^{box}$)} \\
         \cmidrule(lr){2-5}
         & $1.0\times$ & $0.5\times$ & $0.2\times$ & $0.1\times$ \\
         \midrule
          IN-Real~\cite{imagenet} & \textcolor{red}{$39.32$} & $34.97$  & $29.14$ & \underline{$25.51$} \\
         IN-SD1.5~\cite{sd1.5} & $38.89$ & $34.68$ & $28.60$ & $24.05$ \\ 
        IN-GenRobust~\cite{RC} & $38.12$ & $32.11$ & $27.68$ & $23.38$ \\ 
        IN-RealFake~\cite{realfake} &$39.04$ & \underline{$35.09$} & \underline{$29.25$} & $24.88$ \\ 
        \rowcolor{Gray!16}
        IN-lbGen(ours) & \underline{$39.26$} & \textcolor{red}{$35.24$} & \textcolor{red}{$30.68$} & \textcolor{red}{$25.64$} \\
        \bottomrule
        \end{tabular}}
        \label{tab:coco_detect}
    \end{subtable}
    
    \vspace{2mm}

    \begin{subtable}[t]{1.0\linewidth}
        \centering
        \resizebox{0.86\linewidth}{!}{%
        \begin{tabular}{l|cccc}
        \toprule
         \multirow{2}{*}{Pre-Trained Data} & \multicolumn{4}{c}{ADE20K (mIoU)} \\
         \cmidrule(lr){2-5}
         & $1.0\times$ & $0.5\times$ & $0.2\times$ & $0.1\times$ \\
         \midrule
          IN-Real~\cite{imagenet} & \textcolor{red}{$42.44$} & \underline{$38.05$} & $32.10$ & \underline{$27.64$} \\
         IN-SD1.5~\cite{sd1.5} & $41.07$ & $37.62$ & $31.49$ & $26.32$ \\ 
        IN-GenRobust~\cite{RC} & $40.77$ & $37.13$ & $29.36$ & $24.70$ \\ 
        IN-RealFake~\cite{realfake} &\underline{$41.89$} & $37.76$ & \underline{$32.28$} & $27.38$ \\ 
        \rowcolor{Gray!16}
        IN-lbGen(ours) & $41.50$ & \textcolor{red}{$38.61$} & \textcolor{red}{$33.57$} & \textcolor{red}{$27.82$} \\
        \bottomrule
        \end{tabular}}
        \label{tab:ade20k_seg}
    \end{subtable}

    \caption{
        \textbf{Results on COCO object detection and ADE20K semantic segmentation of different number of training images.} We gradually scaling down the number of downstream training images from original data size to $1/10$ of it for testing the generalization ability of the backbones. }
    \label{tab:seg_det}
    \vspace{-1em}
\end{table}

%% file: tabel/specificbias.tex
\begin{table}[t] 
    \centering
    \vspace{0.5em}
    \resizebox{1\linewidth}{!}{\begin{tabular}{llccc}
    \toprule
    Backbone & Pre-Trained Data & $TI$($\downarrow$) & $CB_{avg.}$($\uparrow$) & $BG_{Gap}$($\downarrow$) \\
    \specialrule{2pt}{3pt}{5pt}
    \multirow{5}{*}{ResNet50} & IN-Real~\cite{imagenet} & $60.9$ & $60.0$ & \underline{$6.8$} \\
    & IN-SD1.5~\cite{sd1.5} & \underline{$60.7$} & $55.3$ & $8.0$ \\ 
    & IN-GenRobust~\cite{RC} & $62.8$ & $48.1$ & $7.5$ \\ 
    & IN-RealFake~\cite{realfake} & $69.2$ & \underline{$60.1$} & $8.2$ \\ 
    \rowcolor{Gray!16}
    & IN-lbGen(ours) & \textcolor{red}{$56.1$} & \textcolor{red}{$64.7$} & \textcolor{red}{$6.4$} \\
    
    \specialrule{2pt}{3pt}{5pt}
    \multirow{5}{*}{ViT-S} & IN-Real\cite{imagenet} & $67.0$ & \underline{$61.8$} & \underline{$6.7$} \\
    & IN-SD1.5~\cite{sd1.5} & \underline{$63.8$} & $55.5$ & $7.8$ \\
    & IN-GenRobust~\cite{RC} & $65.7$ & $47.3$ & $7.9$ \\
    & IN-RealFake~\cite{realfake} & $70.6$ & $61.2$ & $7.8$ \\
    \rowcolor{Gray!16}
    & IN-lbGen(ours) & \textcolor{red}{$57.2$} & \textcolor{red}{$66.0$} & \textcolor{red}{$6.1$} \\
    
    \bottomrule
    \end{tabular}}
    \caption{
        \textbf{Results on benchmarks of testing specific bias.} $TI$ (in \%) denotes the texture inclination of the model. $CB_{avg.}$ (in \%) denotes the average relative accuracy when the number of uncommon attributes changes. $BG_{Gap}$ (in \%) metric reports the drop in performance by just changing the background to a different class than the foreground class.}
    \vspace{-1em}
    \label{tab:specific bias}
\end{table}

%% file: tabel/ablation.tex
 \begin{table*}[t] 
    \centering
    \vspace{-1.5em}
    \resizebox{1.0\linewidth}{!}{\begin{tabular}{l|cccc|c|ccccccccc}
    \toprule
    Pre-Trained Data & $\mathcal{L}_{en}$ & $\mathcal{L}_{in}$ & $\mathcal{L}_{q}$ & CLIP Size & IN100-val & Aircraft & Cars196 & DTD & EuroSAT & Flowers & Pets & Food101 & SUN397 & Avg.\\
    \midrule
    IN100-Real~\cite{imagenet100} & -- & -- & -- & -- & $88.3$ & $40.5$ & $28.5$ & $56.8$ & \underline{$92.4$} & \underline{$68.8$} & \underline{$72.1$} & $48.7$ & $38.0$ & $56.0$ \\
    \midrule
    IN100-lbGen & $\checkmark$ & $\checkmark$ & $\checkmark$ & Large & $66.0$ & \textcolor{red}{$44.8$} & \textcolor{red}{$34.3$} & \textcolor{red}{$59.1$} & \textcolor{red}{$92.7$} & \textcolor{red}{$71.6$} & \textcolor{red}{$72.0$} & \textcolor{red}{$50.8$} & \textcolor{red}{$42.3$} & \textcolor{red}{$58.5$} \\
    & $\times$ & $\checkmark$ & $\checkmark$ & Large & $64.6$ & $39.3$ & $27.0$ & $54.8$ & $91.5$ & $65.7$ & $70.3$ & $45.4$ & $35.3$ & $53.6$ \\
     & $\checkmark$ & $\times$ & $\checkmark$ & Large & $8.7$ & $24.5$ & $21.4$ & $45.4$ & $91.4$ & $52.0$ & $43.9$ & $40.1$ & $24.8$ & $42.9$ \\
    & $\checkmark$ & $\checkmark$ & $\times$ & Large & $51.3$ & $41.7$ & $25.1$ & $56.2$ & $92.1$ & $64.4$ & $68.3$ & $46.2$ & $36.4$ & $53.7$ \\
    IN100-SD1.5~\cite{sd1.5} & $\times$ & $\times$ & $\times$ & -- & $65.5$ & $39.7$ & $26.9$ & $53.2$ & $91.1$ & $64.2$ & $69.3$ & $45.6$ & $35.9$ & $53.2$ \\
    \midrule
     & $\checkmark$ & $\checkmark$ & $\checkmark$ & Base & $62.1$ & \underline{$42.3$} & \underline{$32.6$} & \underline{$58.4$} & $92.3$ & $68.2$ & $71.2$ & \underline{$49.5$} & \underline{$39.2$} & \underline{$56.7$}\\
    \bottomrule
    \end{tabular}}
    \caption{
        \textbf{Ablation studies on transfer learning datasets and IN100-val.} Avg. is the average accuracy of eight transfer learning datasets. We also present the results of the model trained on real ImageNet100 for reference.}
    \vspace{-1em}
    \label{tab:ablation}
\end{table*}

%% file: figtex/loss-in.tex
\begin{figure}[t]
    \centering
    \includegraphics[width=\linewidth]{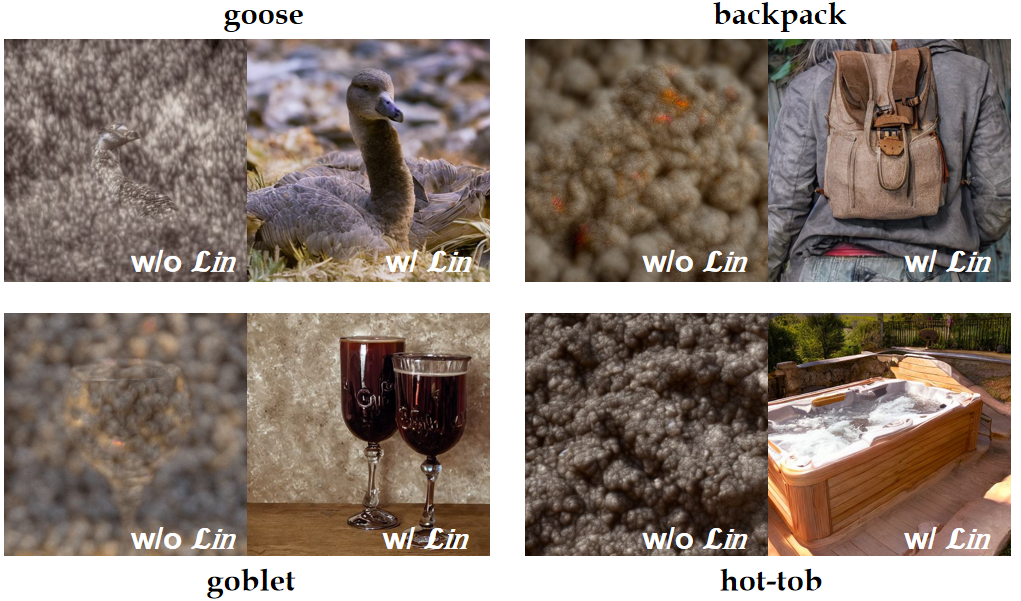}
     \caption{\textbf{Impact of individual image alignmentloss.} We observe that ambiguity problem between classes when discarding $\mathcal{L}_{in}$.}
    \label{fig:loss_in}
     \vspace{-1em}
\end{figure}

%% file: figtex/loss-q.tex
\begin{figure}[t]
    \centering   
    \includegraphics[width=\linewidth]{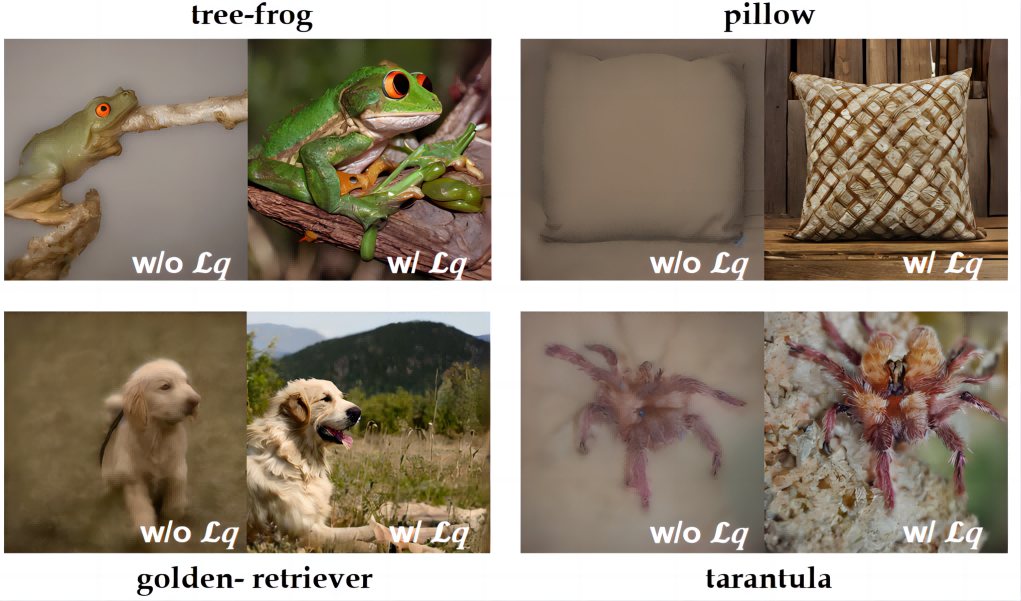}    
    \caption{\textbf{Effectiveness of quality assurance loss.} After adding $\mathcal{L}_{q}$, the image blur problem is solved.}
    \label{fig:loss_q}
    \vspace{-1em}
\end{figure}

%% file: sup/sup_arxiv.tex
\twocolumn[
  \begin{@twocolumnfalse}
    \section*{\centering Appendix of lbGen}
  \end{@twocolumnfalse}
]

\input{tabel/hypset_backbones}
\input{tabel/dataset}
\section{Loss Computation Algorithm}
\label{app:alg}
\begin{algorithm}[h]
  \caption{A complete loss computation step for the lbGen generator during fine-tuning}

\textbf{Input}: class name $c$, semantic description $p_c$, text features of classnames $\{f_{c_1}, \dots ,f_{c_{1000}}\}$, generator $\epsilon_{\theta}$, CLIP model $\mathcal{C}$, discriminator $\mathcal{D}_\phi$, Q-ALIGN model $\mathcal{Q}$, noise $\xi$, scaler $\lambda_1$.
\begin{algorithmic}[1]
\STATE $im = \text{GenerateImage}(\epsilon_{\theta}, \xi, c)$
\STATE $f_{te} = \text{RandomlySelect}(\{f_{c_1}, \dots ,f_{c_{1000}}\})$
\STATE $f_{im}, f_{p_c} = \text{GetFeatures}(\mathcal{C}, im, p_c)$
\STATE $\mathcal{L}_{en}, \mathcal{L}_{neg}  = \text{ComputeEntireLoss}(\mathcal{D}_\phi, f_{im}, f_{te})$
\STATE $\mathcal{L}_{in} = \text{ComputeIndividualLoss}(f_{im}, f_{p_c})$
\STATE $\mathcal{L}_{bi} =  \mathcal{L}_{en} + \mathcal{L}_{in}$
\STATE $\mathcal{L}_{q}, = \text{ComputeQualityLoss}(\mathcal{Q}, im)$
\STATE $\mathcal{L} =  \mathcal{L}_{bi} + \lambda_1 \mathcal{L}_{q}$
\end{algorithmic}
\label{alg:loss}
\textbf{Output}: Training loss for lbGen generator $\mathcal{L}$.
\end{algorithm}

\section{Scoring Quality}
\label{app:score_q}
Q-ALIGN~\cite{qalign} can be recognized as a special version of the multimodal large language model (MLLM). Given an image and system prompt, Q-ALIGN can generate a set of tokens including a {\tt<LEVEL>} token which represents a probability distribution (denoted as $\mathcal{X}$) over all possible tokens. This distribution is then post-processed to derive a score. In the post-processing phase, a closed-set softmax operation is conducted on the set \{$l_i|_{i=1}^{5}\}=\{\textit{bad, poor, fair, good, excellent}\}$ to obtain the probabilities $p_{l_i}$ for each level, such that the sum of $p_{l_i}$ for all $l_i$ equals 1:

\begin{equation}
    p_{l_i} = \frac{e^{\mathcal{X}_{l_i}}}{\sum_{j=1}^{5} {e^{\mathcal{X}_{l_j}}}}.
\end{equation}

As each text level\{\textit{bad, poor, fair, good, excellent}\} corresponds to a score\{\textit{1, 2, 3, 4, 5}\}(higher means better quality), the final predicted score of Q-ALIGN can be formulated as:
\begin{equation}
    \mathcal{S_q} = i \times  \frac{e^{\mathcal{X}_{l_i}}}{\sum_{j=1}^{5} {e^{\mathcal{X}_{l_j}}}},
\end{equation}
where $S_q$ is ranging from one to five.

\section{Training Details} 
\label{app:hyper}
In our fine-tuning method, we inject LoRA layers into the UNet of the diffusion model and train the discriminator from scratch. We keep all other components frozen during training. When training visual backbones, we follow the training recipe in ConvNeXt~\citep{convnext}. It is worth noting that we train Vit-S 40 epochs more than ResNet50 because Transformers often need more time to converge. We provide the detailed training hyperparameters in Table.~\ref{tabel:hyperparam_train} and Table.~\ref{tabel:Backbones_hyperparam}.

What's more, when applying the backbones to downstream tasks, we use the toolbox provided in trex~\cite{trex} to train the linear classifiers for transfer learning. We use MMDetection~\cite{mmdet} and MMSegmentation~\cite{mmseg} toolboxes to train the detection heads and segmentation heads for visual perception tasks, respectively. In the few-shot~\cite{few-shot} setup, we keep the number of training epochs consistent rather than the number of iterations.

\section{Data Synthesis Details}
We use SD1.5~\cite{sd1.5} across all benchmarks. Besides, text prompt ``\texttt{classnames}" and hyperparameters showd in Table~\ref{tab:hyperparameters_gendata} are used to synthesize ImageNet-like datasets (IN-1k, IN-100).
\input{tabel/hypset_gen_data}
\label{app:syn_data}

\section{Datasets Details}
\label{app:datasets}
Except for ImageNet, We also compare with other two synthetic ImageNet datasets~\cite{RC,realfake} because they are the only open source datasets based on SD1.5. Thus, we can get fairer and more convincing results based on one implementation. In addition, all datasets used in our metrics to benchmark the bias of the datasets and test the generalization capacities of the backbones are listed in the Table~\ref{tab:datasets}.
\input{tabel/hypset_gen_train}

\section{Computing Resources}
It takes about 1 hour to fine-tune the generator and 52 hours to generate the ImageNet-like dataset ($\sim$1.3M images) with 8 A100 GPUs. The generation runtime of each image is comparable to existing diffusion models.

\section{Limitation}
While our lbGen demonstrates a great potential to obtain low-biased annotated dataset like ImageNet, the polysemy of some text descriptions may bring drawbacks. As shown in Figure~~\ref{fig:divergent}, some divergences occur when the class name refers to several objects . For instance, the text ``\texttt{crane}" can denote either a bird or a machine, and when prompted with ``\texttt{crane}" to generate a class in our dataset, two entirely different objects will appear.  We consider that these divergences are caused by the multiple directions of clip text space due to the polysemy of human words and may compromise the knowledge of classification models trained on our dataset. Although we believe this issue can be solved with more specific text descriptions instead of class names, how to introduce more specific text descriptions without additional bias other than object is still unclear. We will explore it in our future works.
\input{figtex/divergent}

What's more, our method attempts employing the low-biased text information (e.g., object category name) to regularize and fine-tune the diffusion model in the CLIP feature space for low-biased image generation. Although the diffusion model is only fine-tuned on the 1K categories in ImageNet, our generated dataset shows less bias (i.e., better generalization capacity in downstream tasks) than other competitors. However, on one hand, since the fine-grained categories in ImageNet are scarce, the generalization performance of our method in fine-grained object recognition tasks is still limited. On the other hand, compared with the infinite categories of objects in real world, the number of categories employed for fine-tuning remains limited. This also restrict the generalization capacity of our method, i.e., produces bias. Fortunately, our method provides a general low-biased dataset generation framework, which can mitigate both limitations mentioned above by simply introducing more object categories for fine-tuning.

%% file: tabel/hypset_backbones.tex
\begin{table*}[t]
  \centering
  \caption{Training hyperparameters of \textbf{visual backbones}.
}
  \resizebox{0.88\linewidth}{!}{
  \begin{tabular}{l@{\hspace{1.5cm}}c@{\hspace{1cm}}c@{\hspace{1cm}}c}
    \toprule
    \textbf{Name} & \textbf{ResNet50} & \textbf{ViT-S} & \textbf{ResNet50}(ablation)  \\
    \midrule
    Learning rate & 1e-3 & 1e-3 & 1e-3 \\
    Learning rate scheduler & Cosine decay & Cosine decay & Cosine decay \\
    Epochs & 120 & 160 & 120 \\
    LR warmup epochs & 12 & 16 & 12 \\
    Total batch size & $2048$ & $2048$ & $512$  \\
    Optimizer & AdamW & AdamW & AdamW \\
    AdamW - $\beta_1$ & 0.9 & 0.9 & 0.9 \\
    AdamW - $\beta_2$  & 0.999 & 0.999 & 0.999 \\
    RandAugment & (9, 0.5) & (9, 0.5) & (9, 0.5) \\
    Mixup & 0.8 & 0.8 & 0.8 \\
    CutMix & 1.0 & 1.0 & 1.0 \\
    Random erasing & 0.25 & 0.25 & 0.25 \\
    Label smoothing & 0.1 & 0.1 & 0.1 \\
    Stochastic depth & 0.1/0.4/0.5/0.5 & 0.1/0.4/0.5/0.5 & 0.1/0.4/0.5/0.5 \\
    Layer scale & 1e-6 & 1e-6 & 1e-6 \\
    Head init scale & None & None & None \\
    Gradient clip & None & None & None \\
    Exp. Mov. Avg. (EMA) & 0.9999 & 0.9999 & 0.9999 \\
    \bottomrule
  \end{tabular}
  }

  \label{tabel:Backbones_hyperparam}
\end{table*}

%% file: tabel/dataset.tex
{
\begin{table*}[t]
    \centering
    \small
    \resizebox{0.96\linewidth}{!}{
    \begin{tabular}{@{}lrrrrcc@{}}
    \toprule
    \multicolumn{1}{l}{Dataset} &
        \multicolumn{1}{c}{\# Classes} &
        \multicolumn{1}{c}{\begin{tabular}[c]{@{}c@{}}\# Train\\ samples\end{tabular}} &
        \multicolumn{1}{c}{\begin{tabular}[c]{@{}c@{}}\# Val\\ samples\end{tabular}} &
        \multicolumn{1}{c}{\begin{tabular}[c]{@{}c@{}}\# Test\\ samples\end{tabular}} &
        \multicolumn{1}{c}{\begin{tabular}[c]{@{}c@{}}Val\\ provided\end{tabular}} &
        \multicolumn{1}{c}{\begin{tabular}[c]{@{}c@{}}Test\\ provided\end{tabular}} \\
    \toprule
    \multicolumn{7}{c}{{\em ImageNet validation sets (training classes)}} \\
    ImageNet-Val (IN-val)~\cite{imagenet}    & 1000  & \multicolumn{1}{c}{--} & \multicolumn{1}{c}{--} & 50000            & --  & $\checkmark$ \\
    ImageNet100-Val (IN100-val)~\cite{imagenet100}          & 100  & \multicolumn{1}{c}{--} & \multicolumn{1}{c}{--} & 5000 & --  & $\checkmark$ \\
    \midrule
    \multicolumn{7}{c}{{\em Transfer learning(novel classes)}} \\
    Aircraft~\cite{aircraft}            & 100   & 3334   & 3333  & 3333   & $\checkmark$ & $\checkmark$ \\
    Cars196~\cite{cars}             & 196   & 5700   & 2444  & 8041   & --           & $\checkmark$ \\
    DTD~\cite{dtd}                & 47    & 1880   & 1880  & 1880   & $\checkmark$ & $\checkmark$ \\
    EuroSAT~\cite{eurosat}            & 10    & 13500  & 5400  & 8100   & --           & --           \\
    Flowers~\cite{flowers}          & 102   & 1020   & 1020  & 6149   & $\checkmark$ & $\checkmark$ \\
    Pets~\cite{pets}                  & 37    & 2570   & 1110  & 3669   & --           & $\checkmark$ \\
    Food101~\cite{food101}           & 101   & 68175  & 7575  & 25250  & --           & $\checkmark$ \\
    Sun397~\cite{sun}                 & 397   & 15880  & 3970  & 19850  & --           & $\checkmark$ \\
    \midrule
    \multicolumn{7}{c}{{\em Specific bias (original training classes)}} \\
    Cue Conflict~\cite{cueconf}  & 16  & \multicolumn{1}{c}{--} & \multicolumn{1}{c}{--} & 1280            & --  & $\checkmark$ \\
    FOCUS~\cite{focus} & 226   & \multicolumn{1}{c}{--} & \multicolumn{1}{c}{--} & 23902             & --  & $\checkmark$ \\
    Mixed-Rand \& Mixed-Same~\cite{background} & 9   & \multicolumn{1}{c}{--} & \multicolumn{1}{c}{--} & 8100             & --  & $\checkmark$ \\
    \midrule
    \multicolumn{7}{c}{{\em Visual perception}} \\
    COCO~\cite{coco} & 80 & 118287 & 5000 & 40670 & $\checkmark$ & $\checkmark$ \\
    ADE20K~\cite{ade20k} & 150 & 20210 & 2000 & 3000 & $\checkmark$ & $\checkmark$ \\
    \bottomrule
    \end{tabular}
    }
    \caption{
        {\bf Datasets} we use for evaluating the models.
    }
    \label{tab:datasets}
\end{table*}
}

%% file: tabel/hypset_gen_data.tex
\begin{table}[H]
\centering
\resizebox{\linewidth}{!}{\begin{tabular}{ccccc}
\hline
\textbf{Model} & \textbf{Sampling steps} & \textbf{Scheduler} & \textbf{Guidance scale} & \textbf{Image size} \\
\hline
SD1.5 & 50 & PNDM~\cite{pndm} & 2.0 & \(512 \times 512\) \\
\hline
\end{tabular}}
\caption{Hyperparameters used when synthesizing data.}
\label{tab:hyperparameters_gendata}
\end{table}

%% file: tabel/hypset_gen_train.tex
\begin{table}[t]
  \centering
  \resizebox{1\linewidth}{!}{
  \begin{tabular}{l@{\hspace{1.5cm}}c}
    \toprule
    \textbf{Name} & \textbf{SD1.5} \\
    \midrule
    {\bf Dataset Generator} & \\
    Learning rate & 2e-5 \\
    Learning rate scheduler & Constant \\
    LR warmup steps & 0 \\
    Optimizer & AdamW \\
    AdamW - $\beta_1$ & 0.9 \\
    AdamW - $\beta_2$  & 0.999 \\
    Gradient clipping& $0.1$ \\
    
    \midrule

    {\bf Discriminator} & \\
    Learning rate & 1e-5 \\
    Learning rate scheduler & Constant \\
    Optimizer & AdamW \\
    AdamW - $\beta_1$ & $0$ \\
    AdamW - $\beta_2$  & $0.999$ \\
    Gradient clipping& $1.0$ \\
    
    \midrule
    
    Quality assurance loss weight $\lambda_2$ & 0.1 \\
    Gradient enable steps & $5$ \\
    LoRA rank & 128 \\
    Classifier-free guidance scale & 2 \\
    Resolution & $512 \times 512$ \\
    Total training epochs  & 3 \\
    Local batch size & $4$ \\
    Mixed Precision & FP16 \\
    \bottomrule
  \end{tabular}
  }
  \caption{lbGen training hyperparameters for SD1.5.}
  \label{tabel:hyperparam_train}
\end{table}

%% file: figtex/divergent.tex
\begin{figure}[t]
    \centering

    \includegraphics[width=\linewidth]{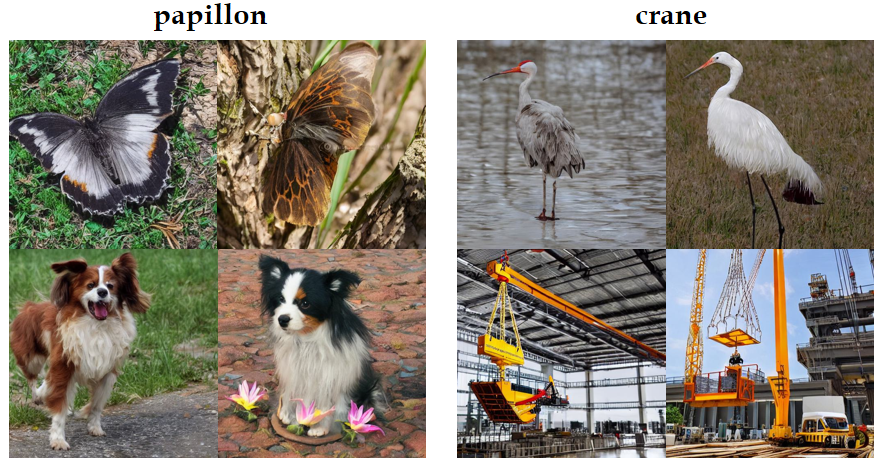}
     \caption{Visualization of generated images prompted by polysemy class name in our dataset.}
    \label{fig:divergent}

\end{figure}